\pdfoutput=1

\documentclass[11pt]{article}

\usepackage{acl}

\usepackage{times}
\usepackage{latexsym}

\usepackage[T1]{fontenc}

\usepackage[utf8]{inputenc}

\usepackage{microtype}

\usepackage{amsthm}
\usepackage{amssymb}
\usepackage{amsmath}
\usepackage{amsfonts}
\usepackage{xcolor}
\usepackage{multirow}
\usepackage{tabularx}
\usepackage{booktabs}
\usepackage{subcaption}
\usepackage{xspace}
\usepackage{graphicx}
\usepackage{color,colortbl}
\usepackage{url}
\usepackage{xspace}
\usepackage{dsfont}
\usepackage{arydshln}
\usepackage{multirow}
\usepackage{flafter}
\usepackage{todonotes}
\makeatletter
\newcommand*\iftodonotes{\if@todonotes@disabled\expandafter\@secondoftwo\else\expandafter\@firstoftwo\fi}
\makeatother

\newcommand{\rparagraph}[1]{\vspace{1.5mm}\noindent\textbf{#1.}}
\newcommand{\sparagraph}[1]{\noindent\textbf{#1.}}

\newcommand{\modelname}{\textsc{QASL}\xspace}
\newcommand{\modelnamep}{\textsc{QASL+}\xspace}

\newcommand{\astfootnote}[1]{
\let\oldthefootnote=\thefootnote
\setcounter{footnote}{0}
\renewcommand{\thefootnote}{\fnsymbol{footnote}}
\footnote{#1}
\let\thefootnote=\oldthefootnote
}
\newcommand\blfootnote[1]{%
  \begingroup
  \renewcommand\thefootnote{}\footnote{#1}%
  \addtocounter{footnote}{-1}%
  \endgroup
}

\title{Improved and Efficient Conversational Slot Labeling \\ through Question Answering}

\author{Gabor Tibor Fuisz$^{*}$, Ivan Vulić, Samuel Gibbons, \\ \textbf{Inigo Casanueva, Paweł Budzianowski}\\
PolyAI Limited, London, UK \\
  \texttt{ivan@poly-ai.com, pawel@poly-ai.com}}

\begin{document}
\maketitle
\begin{abstract}
\renewcommand{\thefootnote}{\fnsymbol{footnote}} 
\blfootnote{$^{*}$Work done during an internship at PolyAI.}
\renewcommand{\thefootnote}{\fnsymbol{footnote}} 
Transformer-based pretrained language models (PLMs) offer unmatched performance across the majority of natural language understanding (NLU) tasks, including a body of question answering (QA) tasks. We hypothesize that improvements in QA methodology can also be directly exploited in dialog NLU; however, dialog tasks must be \textit{reformatted} into QA tasks. In particular, we focus on modeling and studying \textit{slot labeling} (SL), a crucial component of NLU for dialog, through the QA optics, aiming to improve both its performance and efficiency, and make it more effective and resilient to working with limited task data. To this end, we make a series of contributions: \textbf{1)} We demonstrate how QA-tuned PLMs can be applied to the SL task, reaching new state-of-the-art performance, with large gains especially pronounced in such low-data regimes. \textbf{2)} We propose to leverage contextual information, required to tackle ambiguous values, simply through natural language. \textbf{3)} Efficiency and compactness of QA-oriented fine-tuning are boosted through the use of lightweight yet effective adapter modules. \textbf{4)} Trading-off some of the quality of QA datasets for their size, we experiment with larger automatically generated QA datasets for QA-tuning, arriving at even higher performance. Finally, our analysis suggests that our novel QA-based slot labeling models, supported by the PLMs, reach a performance ceiling in high-data regimes, calling for more challenging and more nuanced benchmarks in future work. 


\end{abstract}




\section{Introduction and Motivation}
Task-oriented conversational systems allow users to interact using natural language to solve well-defined tasks such as restaurant booking, hotel assistance, and travel information \citep{young2002talking,raux2005let,Budzianowski:2018emnlp}. \textit{Slot labeling (SL)}, a crucial component of these systems, aims to fill the correct values associated with predefined \textit{slots} from a domain ontology: e.g., a dialog system for hotel reservations is expected to fill slots such as \textit{check in date} and \textit{the number of guests} with the values extracted from a user utterance (e.g., \textit{next Friday}, \textit{4}). However, the manual construction of such domain ontologies and corresponding annotated examples is expensive, time-consuming, and typically requires domain experts as data designers. For this reason, \textit{few-shot} and \textit{data-efficient} SL has drawn a lot of attention recently \citep{hou2020few, henderson2021convex,liu2020coach}, with the aim to maximize data efficiency by learning from only a handful of task-annotated examples. As for the plethora of other NLP tasks \cite{Qiu:2020survey,Razumovskaia:2021survey}, these models typically rely on Transformer-based pretrained language models (PLMs) \citep{devlin2019bert,liu2019roberta}, coupled with SL-specific fine-tuning \cite{henderson2021convex}.


In parallel, machine reading comprehension has been fueled by PLM-based improvements and the creation of large-scale datasets \cite{rajpurkar2018know, fisch2019mrqa}, even matching human-level performance in an array of challenges \citep{devlin2019bert,zhang2020retrospective}. These advances in question answering (QA) models have inspired the ideas of reformatting conversational systems as QA systems \citep{mccann2018natural}. Such QA-reformatting step can be `global' (i.e., it can be applied on the full system), or it can be applied to a particular NLU component, as tried quite extensively for dialog state tracking \citep{gao2019dialog,gao2020machine,zhou2019multi}. 

Recently, \citet{namazifar2021language} have provided preliminary evidence that NLU tasks such as intent detection and slot labeling can also be posed as span-based QA tasks supported by the PLMs: for SL in particular, a question in natural language is defined for each slot, and the answer given by the fine-tuned PLM fills the slot value.\footnote{This formulation is very similar to recent work on prompting task-tuned PLMs \cite{Gao:2021acl}, see also the comprehensive survey on prompting PLMs \cite{Liu:2021survey}.} Performance gains of their QA-based NLU methods, especially in low-data scenarios, indicate the suitability of QA methodology for modeling dialog NLU.

\begin{figure}[!t]
\centering
\vspace{-1.5mm}
\includegraphics[width=0.95\columnwidth]{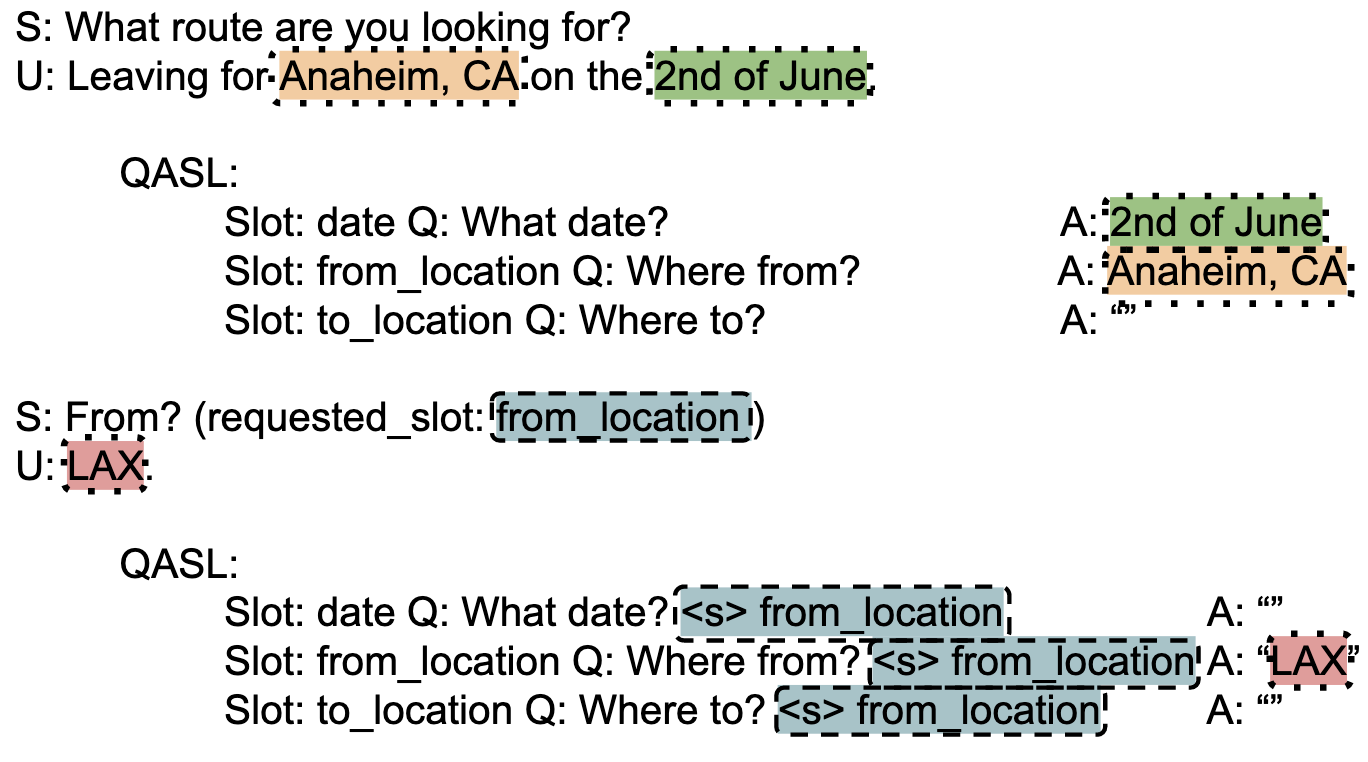}
\caption{Reformulating slot labeling as QA \textit{with} contextual information. S, U, Q, A denote System, User, Question and Answer (dotted lines), respectively. The requested slot (dashed line), indicated in the previous dialog turns, is added to all questions in the current turn. The second example shows a case where contextual information is crucial for slot disambiguation.}
\label{fig:qanlu}
\vspace{-2mm}
\end{figure}

Inspired by this emerging line of research, in this paper we propose the \textbf{\modelname} framework: 
{\bf Q}uestion {\bf A}nswering for {\bf S}lot {\bf L}abeling, which sheds new light on reformatting SL into QA tasks, and studies it extensively from multiple key aspects, while also aiming to align well with `real-world' production-ready settings. We summarize these core aspects as follows:

\vspace{1.0mm}
\noindent \textbf{(1)} The reformulation of SL into QA allows us to benefit from the adaptation of off-the-shelf PLMs and QA-oriented systems to the dialog domain of interest. Are these adaptations robust across domains and datasets, especially for low-data regimes? Further, are they robust with respect to the chosen PLM and the QA dataset selected for QA-based adaptive fine-tuning \cite{Ruder:2021blog}?

\vspace{0.7mm}
\noindent \textbf{(2)} To increase efficiency, current span-based SL models only act over the latest user input; however, in some cases, this simplification deteriorates performance as the context of the conversation is necessary to disambiguate between overlapping slots (see Figure~\ref{fig:qanlu}). How can we adapt \modelname to the inherently contextual nature of dialog while maintaining efficiency? 

\vspace{0.7mm}
\noindent \textbf{(3)} Fully fine-tuning PLMs imposes large training and operational costs, particularly when specialized per-slot SL models are required \cite{namazifar2021language,henderson2021convex,mehri2021gensf}. Is it possible to build more efficient fine-tuning and adaptation approaches? Can such more lightweight QASL models keep up with the performance of full model fine-tuning?


\vspace{0.7mm}
\noindent \textbf{(4)} Can high performance also be obtained with QASL models that leverage larger, automatically generated QA resources for fine-tuning? Can such resources be combined with smaller (but higher-quality) hand-crafted QA resources? 

\vspace{1.0mm}

 In sum, we push further the understanding of key advantages and limitations of the QA-based approach to dialog SL. The proposed \modelname framework is applicable to a wide spectrum of PLMs, and it integrates the contextual information through natural language prompts added to the questions (Figure~\ref{fig:qanlu}). Experiments conducted on standard SL benchmarks and with different QA-based resources demonstrate the usefulness and robustness of \modelname, with state-of-the-art performance, and most prominent gains observed in low-data scenarios. We also verify the viability of artificially created QA resources for the SL task. Finally, we demonstrate that slot-specific SL models can be fine-tuned with less than $1\%$ parameters of the pretrained backbone PLM, while maintaining strong SL performance.
 
 

\section{\modelname: Methodology}
\label{s:methodology}

\sparagraph{Preliminaries}
Following \citet{namazifar2021language}, we pose the SL task as a `pure' question answering problem. 
This reformulation into the QA paradigm maps a list of slots $S$ from the domain ontology to a list of corresponding questions $Q$. For instance, the slots \emph{date},  \emph{from\_location}, \emph{to\_location}, 
can be posed as simple natural questions as follows: \textit{``What date?''}, \textit{``Where from?''}, \textit{``Where to?''}, respectively; see Figure~\ref{fig:qanlu}.\footnote{The mapping between S and Q can be one-to-many.} At each dialog turn, given the input context $C$, which may comprise one or more previous turns, the model is sequentially queried with all pre-defined questions appended to $C$, and returns an answer as a span extracted from the input user utterance, see Figure~\ref{fig:qanlu} again.

\rparagraph{Fine-Tuning Stages in a Nutshell}
\label{sec:stages}
We start from any standard Transformer-based \cite{vaswani2017attention} PLM such as BERT \cite{devlin2019bert}, RoBERTa \cite{liu2019roberta}, or ELECTRA \cite{clark2020electra}. 
Next, in Stage 1 termed \emph{QA-tuning}, the underlying PLM is fine-tuned with a span-based QA objective using large QA datasets such as SQuAD \citep{rajpurkar2018know} or MRQA \citep{fisch2019mrqa}. The goal of Stage 1 is to adapt the model to the span extraction task \cite{Ruder:2021blog} with (large and general-purpose) QA data, and this way effectively increase the model's ability to cope with many different questions. Following that, in Stage 2 termed \emph{QASL-tuning}, the model is fine-tuned further for a particular dialog domain. In this stage, the model further specializes to the small subset of in-domain questions that correspond to the slots from the domain ontology.

\subsection{\modelname with Contextual Information}
\label{s:contextual}
In complex domains with multiple slots, values can often overlap, which might result in severe prediction ambiguities.\footnote{For instance, in the domain of restaurant booking, values for the slots \emph{time} and \emph{people} can both be answered with a single number (e.g., \textit{6}) as the only information in the user utterance, causing ambiguity. In another example, Figure~\ref{fig:qanlu} shows a conversation from the Buses domain in the DSTC8 dataset \cite{rastogi2020schema}; here, it is impossible to distinguish between \emph{from\_location} and \emph{to\_location} without context.} The correct prediction can be only made given the context of the conversation.  

Moreover, natural conversations are of mixed initiative, where the user can provide more information than it was requested or unexpectedly change the dialog topic \citep{rastogi2020schema}. Carrying over the contextual knowledge is a fundamental feature of a successful dialog system \citep{heck2020trippy}. However, a standard straightforward approach, adopted by the current span-based SL models \cite{henderson2021convex,namazifar2021language} to boost simplicity and efficiency, runs inference only over the latest user utterance without context or reserves extra parameters for the slots that have been explicitly requested by the system. Put simply, many current approaches discard the potentially crucial contextual information.

In practice, some contextual information from previous dialog turns can be formulated into the so-called \textit{requested slot} \cite{coope2020span}: this means that the current dialog turn is additionally annotated with the slots requested by the system \cite{coope2020span,rastogi2020schema}, helping slot disambiguation. We propose to provide that information to \modelname by simply appending the requested slot feature (as a natural language prompt) to the posed question, without any architectural modification, as illustrated in Figure~\ref{fig:qanlu}. For instance, if the requested slot is present for the slot \emph{arrival\_time}, and the current question concerns the slot \emph{date}, the final question takes the following form: \textit{``What dates are you looking for \textnormal{\texttt{<s>}} arrival time''}, where \texttt{<s>} is a special separator token. When multiple slots are requested, they are all appended to the initial question, each slot separated by one separator token \texttt{<s>}.

\begin{figure}[!t]
\centering
 \includegraphics[width=.75\columnwidth]{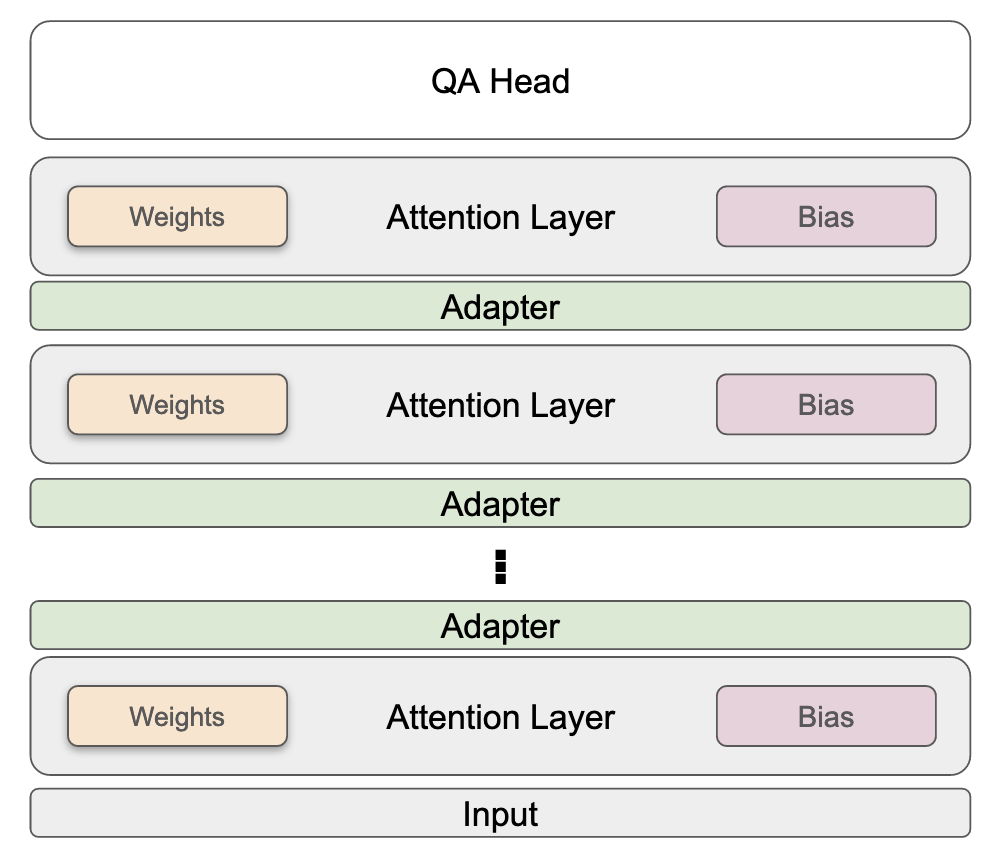}
 \vspace{-1mm}
\caption{A compact illustration of different efficient fine-tuning schemes used with \modelname, and their corresponding parameter subsets (see \S\ref{s:efficiency}).}
\label{fig:fine_tuning}
\vspace{-2mm}
\end{figure}

\subsection{Refining QA-Tuning}
\label{s:qatuning}
Stage 1 of \modelname \textit{QA-tuning} is concerned with adaptive transformations of the input PLMs to (general-purpose) span extractors, before the final in-task \textit{QASL-tuning}. We also propose to further refine Stage 1 and divide it into two sub-stages: (a) \textit{Stage 1a} then focuses on fine-tuning on larger but noisier, automatically generated QA datasets, such as PAQ \cite{lewis2021paq}; (b) \textit{Stage 1b} continues on the output of Stage 1a, but leverages smaller, manually created and thus higher-quality QA datasets such as SQuAD$2.0$ \cite{rajpurkar2018know} and/or MRQA \cite{fisch2019mrqa}.

The rationale behind this refined multi-step \textit{QA-tuning} procedure is that the models 1) should leverage large quantities of automatically generated (QA) data and a task objective aligned with the final task \cite{henderson2021convex}, that is, large-scale adaptive fine-tuning \cite{Ruder:2021blog} before 2) getting `polished' (i.e., further specialized towards the final task) leveraging fewer high-quality data. We refer to the \modelname model variants which rely on the refined Stage 1 procedure as \modelnamep.

\subsection{Efficient \modelname}
\label{s:efficiency}
In principle, one model could be employed to serve all slots in all domains across different deployments. This, however, prevents the separation of different data sources of data, while this is often required from the perspective of data privacy. On the other hand, storing separate slot-specific and domain-specific models derived from heavily parameterized PLMs is extremely storage-inefficient, and their fine-tuning can be prohibitively slow \cite{henderson2021convex}.\footnote{Distilling PLMs to their smaller counterparts \cite{lan2019albert,sanh2019distilbert} does not resolve the issue for production-oriented deployments.} Therefore, with multiple domains and slots, the model compactness and fine-tuning efficiency become crucial features. In order to address these requirements, we rely on and experiment with three different efficiency- and compactness-oriented approaches within the \modelname framework in Stage 2, also summarized in Figure~\ref{fig:fine_tuning}:

\vspace{0.8mm}
\noindent \textbf{(1)} Fine-tuning only the QASL model's \textit{head}, which is responsible for predicting the start and the end of the answer span. All other parameters are kept fixed/frozen. Most QA systems based on PLMs contain a simple one feed-forward layer as the head, using $\leq0.1\%$ of all the parameters.

\vspace{0.5mm}
\noindent \textbf{(2)} Using lightweight tunable bottleneck layers, that is, \textit{adapters} \cite{houlsby2019parameter,pfeiffer2020adapterfusion}, inserted within each Transformer layer of the underlying model. At fine-tuning, only adapter parameters are updated while all the other parameters of the model are kept fixed: i.e., typically $\leq1\%$ of the PLM's original parameter capacity gets updated \cite{pfeiffer2020adapterfusion}. 

\vspace{0.5mm}
\noindent \textbf{(3)} Fine-tuning only \textit{bias} parameters of the attention layers: this approach, termed BitFit \cite{zaken2021bitfit} in practice fine-tunes less than $0.1\%$ of the full parameters.

\vspace{0.8mm}

It is worth noting that adapters and bias-only tuning (i.e., BitFit) have been evaluated only in full task-data setups in prior work. Here, our use-case scenario adds another layer of complexity as we evaluate them in few-shot scenarios of the SL task.


\section{Experimental Setup}
\label{sec:experiments}

\sparagraph{Underlying PLMs}
We opt for a set of established PLMs with strong performance record on other NLP tasks: RoBERTa \cite{liu2019roberta} (its Base and Large variants), and a distilled version of BERT -- DistilBERT \cite{sanh2019distilbert}. However, we note that \modelname is applicable also to other PLMs.\footnote{For instance, we have run experiments also with ELECTRA \cite{clark2020electra}, but do not report its performance as it was consistently outperformed by RoBERTa.}




\rparagraph{QA Datasets (Stage 1)}
We experiment with two manually created QA datasets, (i) SQuAD$2.0$ \cite{rajpurkar2018know}, and (ii) MRQA \cite{fisch2019mrqa}; and (iii) one automatically generated QA dataset, PAQ \cite{lewis2021paq}. SQuAD$2.0$ was also used in prior work of \citet{namazifar2021language}: it consists of $150$k QA pairs including $50$k negative pairs without any answer. The MRQA dataset is a collection of $18$ existing QA datasets, spanning almost $2$M QA pairs, converted to the same format of SQuAD$2.0$. The PAQ dataset, created for open-domain QA, consists of over $65$M natural language QA pairs. Due to hardware constraints, we randomly sample two smaller versions from the full PAQ, spanning 5M and 20M QA pairs and denoted as PAQ$5$ and PAQ$20$; they are also adapted to the same SQuAD$2.0$ format.

By selecting these diverse QA-data sources, we validate and compare their usefulness for adaptive QA fine-tuning oriented towards SL, reaching beyond SQuAD$2.0$ as a standard go-to dataset. We also test if the sheer scale of an automatically generated dataset (i.e., PAQ) can compensate for its lower data quality, compared to manually created SQuAD and MRQA.

\rparagraph{Slot Labeling Datasets: Stage 2 and Evaluation}
We run experiments on two standard and commonly used SL benchmarks: (i) RESTAURANTS-$8$k \citep{coope2020span} and DSTC8 \citep{rastogi2020schema}, which are covered by the established DialoGLUE benchmark \cite{mehri2020dialoglue}.

RESTAURANTS-$8$k comprises conversations from a commercial restaurant booking system, and covers 5 slots required for the booking task: \textit{date}, \textit{time}, \textit{people}, \textit{first name}, and \textit{last name}, with a total of 8,198 examples over all 5 slots, see the work of \citet{coope2020span} for further details.

DSTC8 has been introduced during the Dialog System Technology Challenge (DSTC) 8 challenge, and then adapted to the span extraction task by \citet{coope2020span}. It includes over 20k annotated multi-domain, task-oriented conversations between humans and a virtual assistant. These conversations involve interactions with services and APIs spanning $4$ domains (Buses, Rental Cars, Events, and Homes) and $12$ slots; see \citet{rastogi2020schema}.

Similar to prior work \cite{coope2020span,henderson2021convex,mehri2021gensf}, we also do tests where we fine-tune on smaller \textit{few-shot data samples} of the two SL datasets, while always evaluating on the same (full) test set. RESTAURANTS-$8$k comes with $8$ different few-shot data samples referred to as $1/128$, $1/64$, $1/32$, $1/16$, $1/8$, $1/4$, $1/2$, $1$ (proportions of the full dataset). Similarly, we fine-tune on $1/32$, $1/16$, $1/8$, $1/4$, $1/2$, $1$ proportions of the full DSTC8.\footnote{The exact numbers are in Appendix~\ref{sec:appendix} (Table~\ref{tab:data_splits}).}

\rparagraph{\modelname: Fine-tuning Setup and Hyperparameters}
Our \modelname implementation is based on the Transformers library \citep{wolf2019huggingface}. Each PLM is equipped with a QA-head which is a feed-forward network with two outputs to compute span start logits and span end logits. 

Stage 1 is carried out on $8$ V100 GPUs for $2$ epochs with $24$ QA-pairs per batch per GPU, relying on the Adam optimizer \citep{kingma2014adam} with a learning rate of $3$e-$5$. We investigate the following 8 Stage 1 (i.e., QA-tuning) regimes: SQuAD, MRQA, PAQ5, PAQ20 (basic \modelname), PAQ5-SQuAD, PAQ5-MRQA, PAQ20-SQuAD, and PAQ20-MRQA (\modelnamep, see \S\ref{s:qatuning}). The basic Stage 1 setup, unless noted otherwise, is QA-tuning on SQuAD.

Stage 2 (\modelname-tuning) proceeds in batches of size $32$, again with Adam, and a learning rate $2$e-$5$. All presented results are averaged over $5$ different runs. We follow the setup from prior work \citep{coope2020span,henderson2021convex,mehri2021gensf}, where all the hyper-parameters are fixed across all domains and slots. The reported evaluation metric is the average F1 score across all slots in a given task/domain.\footnote{It is computed with an exact score, that is, the model has to extract exactly the same span as the golden annotation. This is different to a typical QA setup where partial or multiple answers are also taken into account \citep{rajpurkar2018know}.}

\rparagraph{Baselines} We compare \modelname against three recent state-of-the-art SL models:\footnote{For full technical details of each baseline model, we refer the reader to their respective papers.} 

\vspace{0.8mm}
\noindent \textbf{ConVEx} \citep{henderson2021convex} defines a novel SL-oriented pretraining objective, termed \textit{pairwise sentence cloze}, combined with SL-tuning of only a subset of parameters. It shows strong performance particularly in few-shot scenarios.

\vspace{0.6mm}
\noindent \textbf{GenSF} \citep{mehri2021gensf} adapts the pretrained DialoGPT model \citep{zhang2020dialogpt} and  steers/constrains its generation freedom to reflect the particular dialog domain; at the same time it adapts the downstream SL task to align better with the architecture of the (fine-tuned) DialoGPT.

\vspace{0.6mm}
\noindent \textbf{QANLU} \citep{namazifar2021language} also reformulates SL as a QA task (see \S\ref{s:methodology}) by performing in-task fine-tuning of DistilBERT$_{Base}$ model \cite{sanh2019distilbert} which was first fine-tuned on SQuAD$2.0$. 

\begin{figure}[!t]
\centering
\includegraphics[width=0.9\columnwidth]{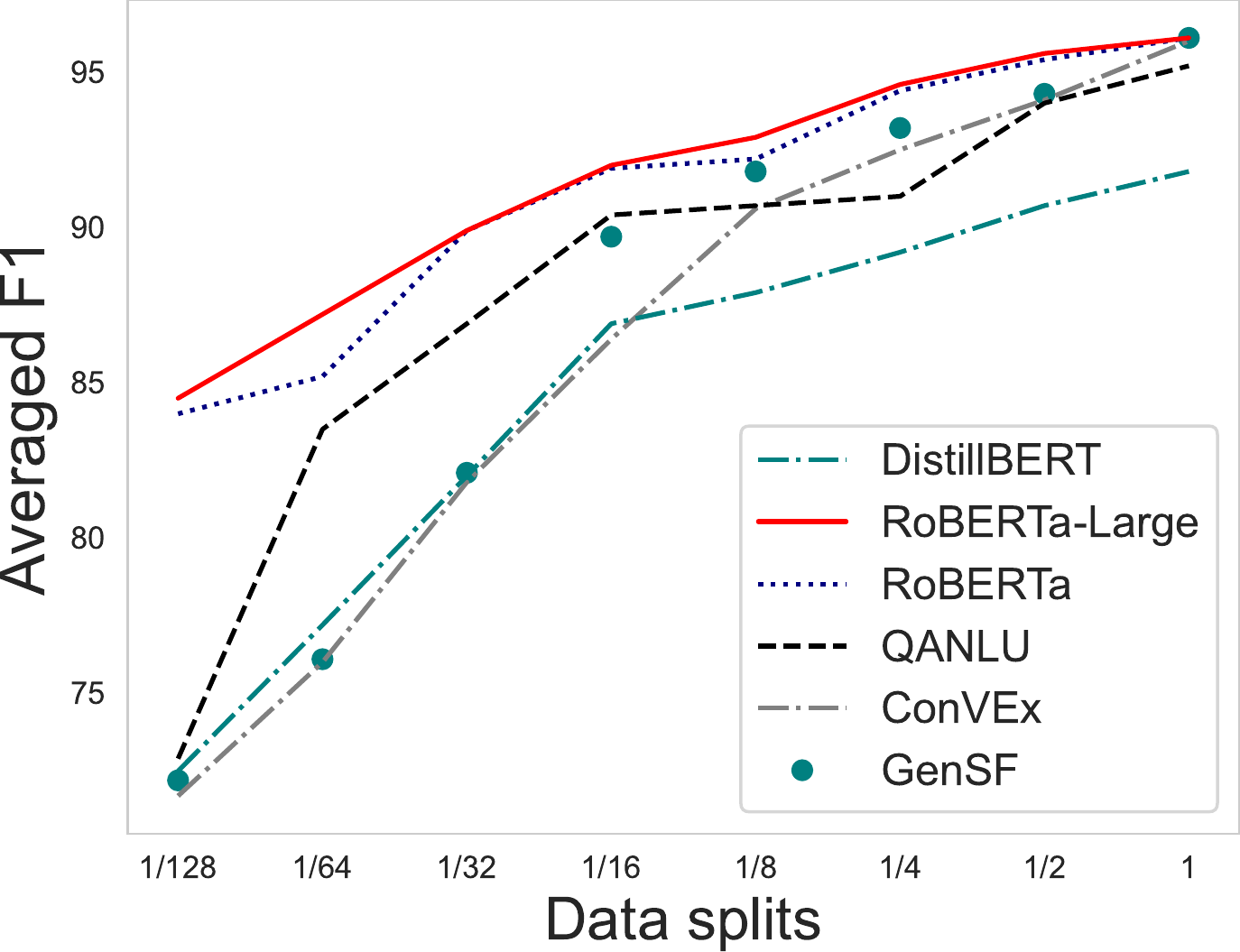}
\vspace{-1mm}
\caption{A comparison of slot labeling models on RESTAURANTS-8k. Stage 1 for \modelname and QANLU are run on SQuAD2.0. x-axis shows the fraction of the training data used for SL-tuning (see \S\ref{sec:experiments}).}
\label{fig:r8k_default}
\vspace{-1.5mm}
\end{figure}
\vspace{0.5mm}
\rparagraph{Efficient \modelname in Stage 2: Setup}
For QA-head-only tuning, the default head in Transformers library \citep{wolf2019huggingface} is a linear layer of size $[E,2]$ where $E$ is the size of the output embedding of the PLM. We define the QA head as a feed-forward network with $2$ layers, covering $\approx1$M parameter. For experiments with adapters, we rely on the lightweight yet effective Pfeiffer architecture \citep{pfeiffer2020adapterfusion}, using the reduction factor of $16$ for all but the first and last Transformers layer, where the factor of $8$ was utilized.\footnote{The learning rate has been increased to $1$e$-3$ following prior work \citep{pfeiffer2020adapterfusion}, and it also yielded better performance in our preliminary experiments.}


\section{Results and Discussion}

\sparagraph{\modelname versus Baselines}
In the first experiment, we benchmark \modelname against all baseline models and across different levels of data availability for Stage 2 SL-tuning. We assume SQuAD$2.0$ as the underlying QA dataset for Stage 1 for all models (including the baseline QANLU), and do not integrate contextual information here (see \S\ref{s:contextual}). Figure~\ref{fig:r8k_default} plots the results on RESTAURANTS-$8$k,\footnote{The exact numbers are in Appendix~\ref{sec:appendix} (Table~\ref{tab:r8k}).} and reveals several findings. First, there is an indication that larger models yield performance gains: RoBERTa$_{Large}$ is slightly stronger than RoBERTa${_{Base}}$ as the underlying model, although RoBERTa${_{Base}}$ also shows very competitive performance across the board. While most models reach very similar and very high performance in the full-data regime, the difference between models becomes much more salient in few-shot setups. The gains in favor of \modelname with RoBERTa-s over all baselines are the largest for the scarcest data scenarios: $1/64$ and $1/128$.\footnote{While we also achieve higher results than QANLU \citep{namazifar2021language}, the exact comparison is not possible since they used different data splits.} \footnote{We have also tried different question prompts but have not observed any significant variance in the results.}

\rparagraph{Using Contextual information}
We now investigate if the integration of contextual information in the form of requested slots improves SL performance (see \S\ref{s:contextual}). Unless noted otherwise, from now on we assume that \modelname always integrates the requested slot information. The results on RESTAURANTS-8k for a subset of test examples with non-empty requested slots (i.e., 897 out of all 3,731 test examples), are summarized in Table~\ref{tab:r8k_req}. The variant with requested slot information consistently yields higher $F_1$ scores, even despite the fact that the test set contains only $86$ examples that might cause ambiguity. 

\begin{figure*}[t!]

\begin{subfigure}[b]{0.38\linewidth}
\centering
\includegraphics[width=\linewidth]{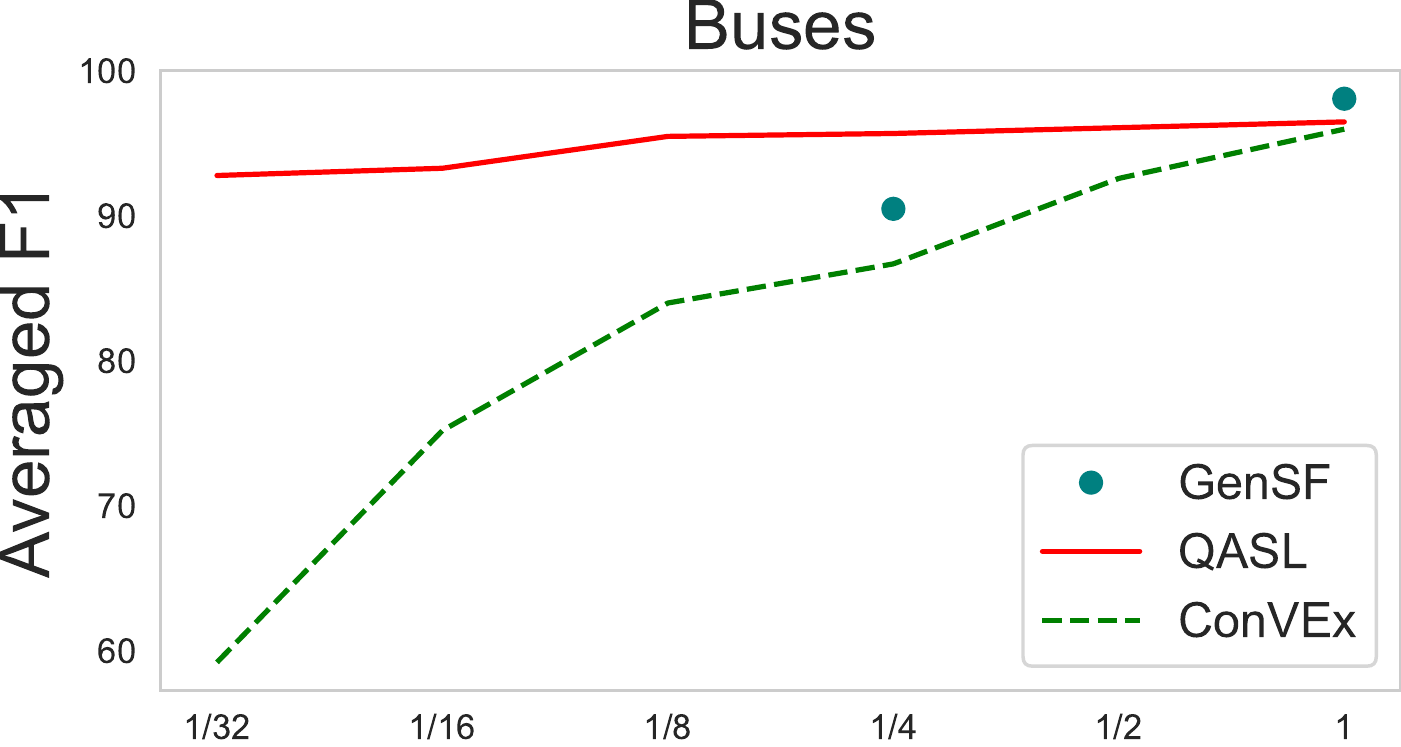}
\end{subfigure}
\hfill
\begin{subfigure}[b]{0.38\textwidth}
\centering
\includegraphics[width=\linewidth]{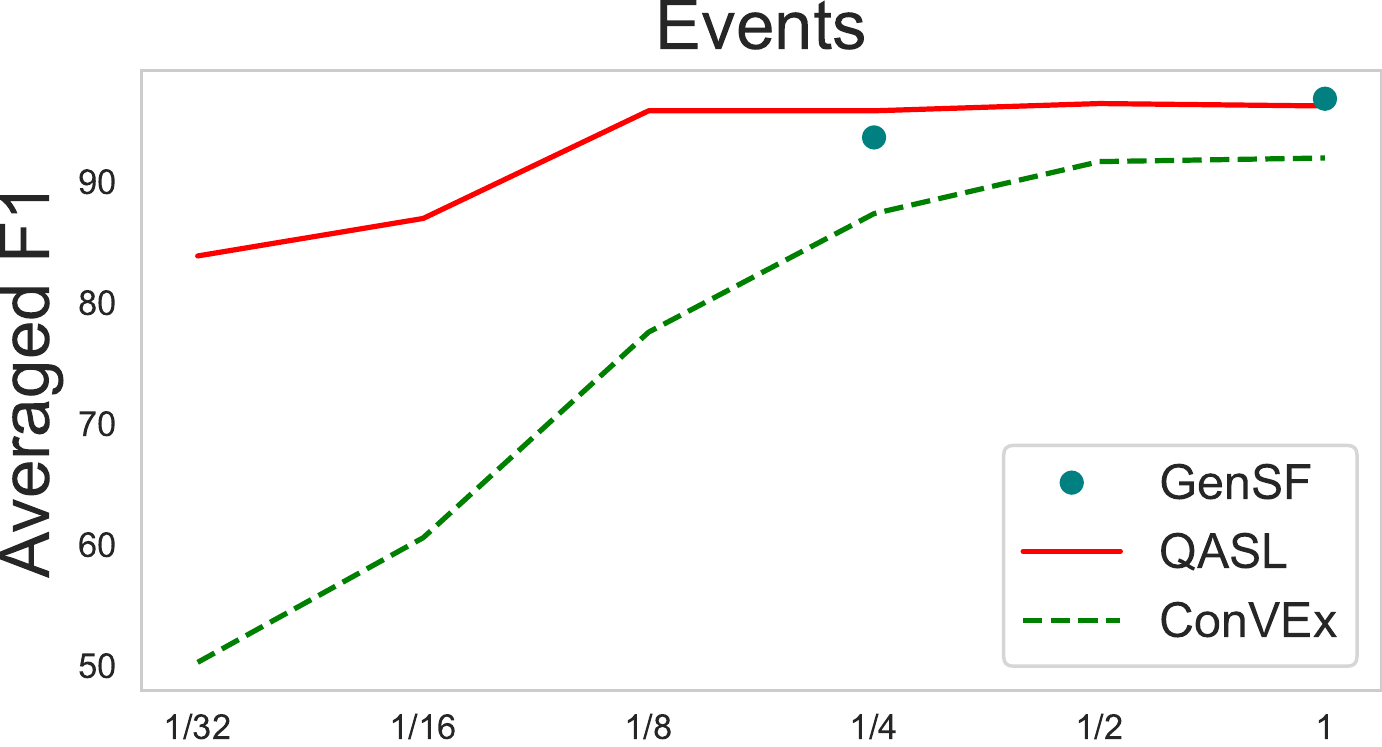}
\end{subfigure}
\begin{subfigure}[b]{0.38\linewidth}
\includegraphics[width=\linewidth]{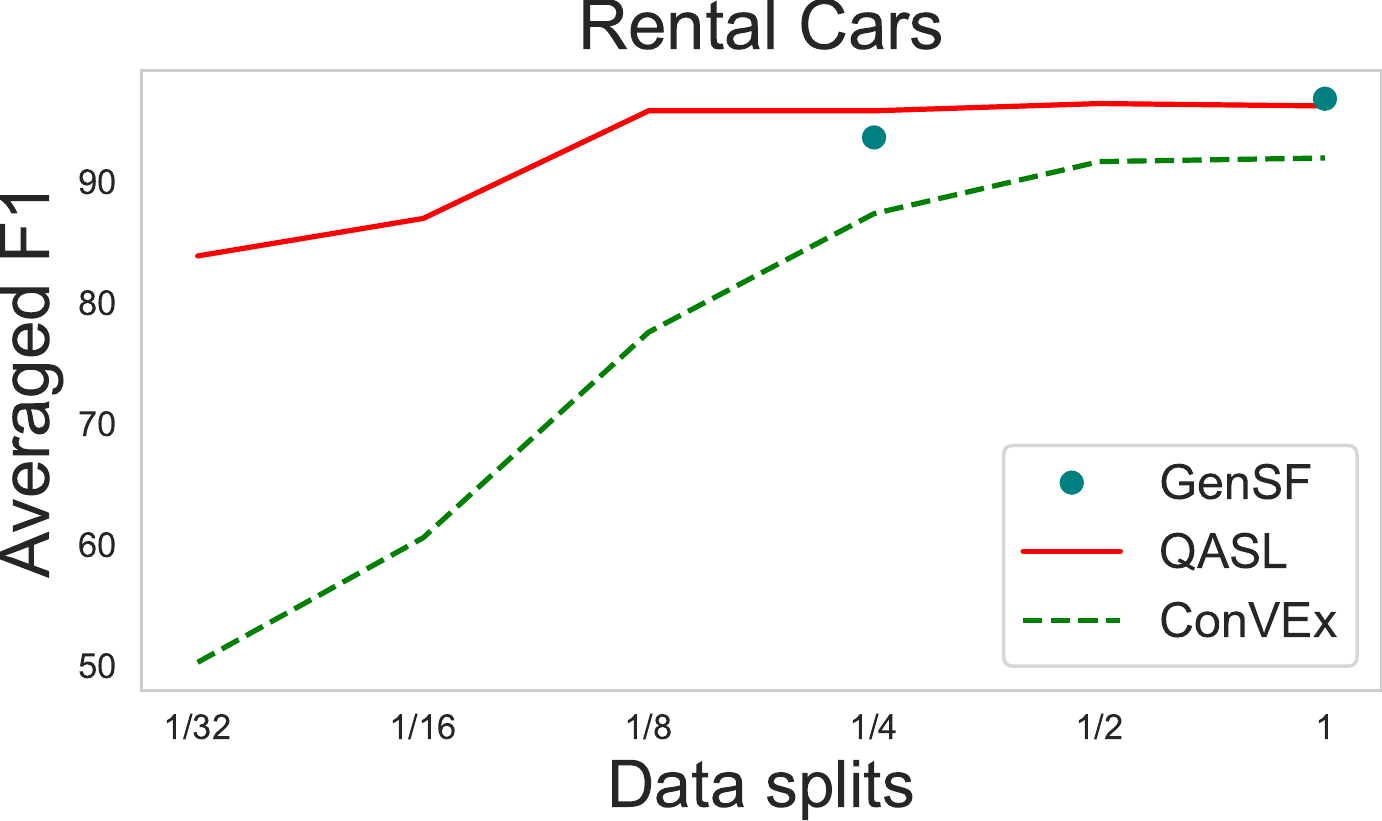}
\end{subfigure}
\hfill
\begin{subfigure}[b]{0.38\textwidth}
\includegraphics[width=\linewidth]{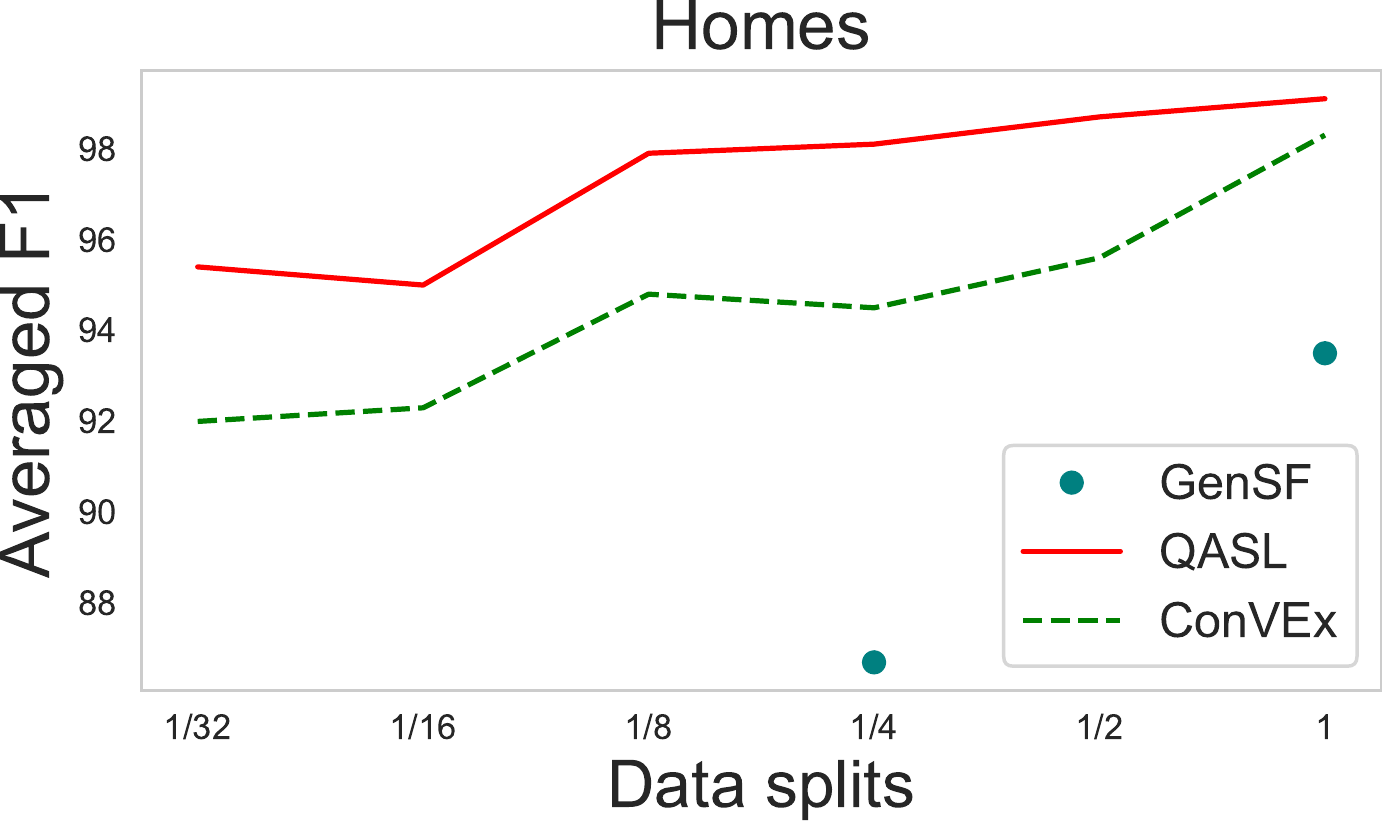}
\end{subfigure}
\caption{Results on the DSTC$8$ dataset across $4$ domains. The performance of GenSF is taken from the original paper and is only available for two data splits: 1 (full data) and $1/4$. The \modelname fine-tunes RoBERTa$_{Large}$ on SQUAD2.0 in Stage 1, and uses contextual requested slot information in Stage 2.}
\label{fig:dstc8}
\end{figure*}

The results on the 4 domains of DSTC8, provided in Figure~\ref{fig:dstc8} for all test examples, show very similar patterns and improvements over the baseline SL models GenSF and ConVEx, especially in few-shot scenarios. The gains with the contextual variant are less pronounced than in RESTAURANTS-8k as DSTC8 covers a fewer number of ambiguous test examples.


\begin{table}[!t]
{\footnotesize
\def\arraystretch{0.76}
\centering
\begin{tabularx}{\linewidth}{l XX}
\toprule 
& \multicolumn{1}{l}{\bf Without Requested} & \multicolumn{1}{l}{\bf With Requested}  \\
\midrule 
1/128 & 81.7 & \textbf{85.8}\\
1/64  & 81.0 & \textbf{87.9}\\      
1/32  & 86.7 & \textbf{90.7}\\      
1/16  & 88.7 &\textbf{93.8}\\   
1/8  & 88.9  &\textbf{95.7}  \\   
1/4  & 91.0 &\textbf{95.0}     \\   
1/2  & 91.5 &\textbf{97.0}\\   
1  &  92.0 & \textbf{98.0}\\   
\bottomrule
\end{tabularx}
}
\vspace{-1mm}
\caption{A comparison of \modelname without and with requested slot information on the subset of RESTAURANTS-8k test examples with non-empty requested slots (891 test examples).}
\label{tab:r8k_req}
\vspace{-1.5mm}
\end{table}



Further, we observe extremely high absolute scores, especially in higher-data setups, which is the first indication that the standard SL benchmarks might become inadequate to distinguish between SL models in the future. We provide a finer-grained analysis of the SL benchmarks later in \S\ref{sec:data}.

\rparagraph{Efficient Fine-Tuning in Stage 2}
We now proceed with the RoBERTa$_{Base}$ model as our base PLM in all following experiments: it achieves very competitive results while using $\approx$3 times fewer parameters than RoBERTa$_{Large}$. Table~\ref{tab:adapters} presents the scores obtained with the three efficient fine-tuning approaches (see \S\ref{s:efficiency}) on RESTAURANTS-8k in few-shot scenarios. 
\begin{table}[h!]
{\footnotesize
\def\arraystretch{0.76}
\centering
\begin{tabularx}{\linewidth}{l X XXX}
\toprule 
& \multicolumn{1}{l}{\bf Full} & \multicolumn{1}{l}{\bf QA head} & \multicolumn{1}{l}{\bf BitFit} & \multicolumn{1}{l}{\bf Adapters}  \\
\cmidrule(lr){2-2} \cmidrule(lr){3-5}
1/128 &  \textbf{84.0} & 0.0 & 25.6 & 81.9 \\
1/64  & \textbf{85.2} & 20.2 & 27.9  & 85.0 \\      
1/32  & 89.9 & 28.3 & 32.5 & \textbf{91.0}  \\      
1/16  & 91.9 & 33.8 & 52.0  & \textbf{92.9} \\   
1/8  & 92.2 & 40.7 & 51.7 & \textbf{93.6}      \\   
1/4  & 94.4 & 52.6 & 70.5 & \textbf{95.2}      \\   
1/2  & 95.4 & 57.7 & 88.8  & \textbf{96.1}  \\   
1  & 96.1 & 61.8 & 93.6  & \textbf{97.0}   \\ 
\bottomrule
\end{tabularx}
}
\caption{Average $F_1$ scores across all slots on the entire RESTAURANTS-8k test data with efficient fine-tuning architectures in Stage 2 (see \S\ref{s:efficiency}), and their comparison to \textbf{Full} model fine-tuning.}
\label{tab:adapters}
\vspace{-1.5mm}
\end{table}

Overall, the results indicate that few-shot scenarios are quite challenging for efficient fine-tuning methods, typically evaluated only in full-data scenarios in prior work \cite{zaken2021bitfit}. The adapter-based approach is most effective by far, and is very competitive to full model fine-tuning, even outperforming it in all but the two fewest-data scenarios. The other two efficient approaches fall largely behind in all training setups. In summary, the results empirically validate that adapter-based fine-tuning offers a viable trade-off between performance and efficiency, even in low-data regimes: it fine-tunes only $\approx$1.5M parameters, translating to $5$MB of storage space, compared to 110M parameters (i.e., 550 MB) needed for full fine-tuning.

\begin{table*}[t!]
\footnotesize
\def\arraystretch{0.76}
\centering
\begin{tabularx}{\linewidth}{l XXXXXXXX}
\toprule 
& \multicolumn{1}{l}{\bf SQuAD} & \multicolumn{1}{l}{\bf MRQA} & \multicolumn{1}{l}{\bf PAQ5} & \multicolumn{1}{l}{\bf PAQ20} & \multicolumn{1}{l}{\bf PAQ5-SQuAD} & \multicolumn{1}{l}{\bf PAQ20-SQuAD} &
\multicolumn{1}{l}{\bf PAQ5-MRQA} & \multicolumn{1}{l}{\bf PAQ20-MRQA}\\
\midrule 
1/128 & 84.0 &	\textbf{86.31} &83.62&	82.57&	    86.09&	85.19&  \textbf{86.31}  &	85.47\\
1/64  & 85.2 &	87.59 &         86.45    &	85.64 &	87.95&	87.11&	\textbf{88.40} &	87.65 \\
1/32 &  89.9 &	\textbf{91.50 }& 91.14 & 89.97 &	   91.46 &	90.92&	91.13 &	91.08 \\ 
\bottomrule
\end{tabularx}
\vspace{-1mm}
\caption{$F_1$ scores over all slots on the RESTAURANTS-$8$k test data for different QA-tuning regimes in Stage 1.}
\label{tab:paq_mrqa}
\vspace{-1.5mm}
\end{table*}

\rparagraph{Different Stage 1 Fine-Tuning Schemes}
Note that, until now, the results were based solely on models QA-tuned with SQuAD$2.0$ in Stage 1. We now test the impact of the QA resource in Stage 1 on the final SL performance. Table~\ref{tab:paq_mrqa} presents the results for the $8$ Stage 1 regimes (see \S\ref{sec:experiments}), fine-tuned with \modelname on $3$ smallest RESTAURANTS-$8$k training data splits in Stage 2.

When using only one QA dataset in Stage 1, several trends emerge. First, a larger of the two manually created datasets, MRQA, yields consistent gains over SQuAD2.0, over all training data splits. Using larger but automatically created PAQ5 and PAQ20 is on par or even better than using SQuAD, but they cannot match performance with MRQA. This confirms that both QA dataset quality \textit{and} dataset size play an important role in the two-stage adaptation of PLMs into effective slot labellers. Having more PAQ data typically yields worse performance: it seems that more noise from more automatically generated QA pairs gets inserted into the fine-tuning process (cf., PAQ20 versus PAQ5). 

However, \modelname tuned only with automatically generated data is still on par or better than tuning with SQuAD$2.0$. This proves the potential of large-scale (automatically obtained) QA datasets for QA-based slot-labeling in domains that have a small overlap with curated QA data such as SQuAD. The highest gains over SQuAD when using PAQ are obtained for two slots: $first\_name$ and $last\_name$. This stems from the fact that finding the right person's name is a common task with Wikipedia-related corpora. Finally, in two out of the three training data splits, the peak scores are achieved with the refined Stage 1 (the PAQ5-MRQA variant), but the gains of the more expensive PAQ5-MRQA regime over MRQA are mostly inconsequential. 


\section{SL Data Analysis and Audit}
\label{sec:data}

Detected high absolute scores in full-data setups for many models in our comparison (e.g., see Figure~\ref{fig:r8k_default}, Table~\ref{tab:adapters}, Figure~\ref{fig:dstc8}) suggest that the current SL benchmarks might not be able to distinguish between state-of-the-art SL models. The remaining gap to 100\% performance might also be due to annotation errors and inconsistencies. We thus inspect the two SL benchmarks in more detail.


On RESTAURANTS-8k, we found that adding the contextual information robustly resolves the issue of ambiguous one-word utterance examples. We identified $86$ examples where the utterance is a single number, intentionally meant to test the model’s capability of using the requested slot, as they could refer either to \textit{time} or \textit{number of people}. Adding requested slot information eliminates all but $2$ of these mistakes. Another challenging group of example concerns {rare} names - most of the issues come from mixing up \textit{first name} and \textit{last name} since both are requested together.

Upon inspection of RESTAURANTS-8k's test set, we discovered several annotation issues. Analyzed models perform the worst on the \textit{time} slot. This is partly due to the many ways one can express time, but also owning to difficulties in annotations. In the test set, some time examples are in the format \textit{TIME pm}, while others use \textit{TIME p.m.}: in simple words, whether the \textit{pm} postfix is annotated or not is inconsistent. Another inconsistency concerns preposition annotations such as \textit{on}, \textit{at}. In some examples the prepositions are included in the answer (e.g. \textit{is there a table free at 8 in the morning}), in others they are not. A similar challenge concerns annotating `the' in \textit{date} answers, such as \textit{the first Sunday of September} instead of \textit{first Sunday of September}. This leads the model to select \textit{August 23rd} instead of \textit{the day of August 23rd}.
Another annotation inconsistency concerns the \textit{people} slot. In some examples, only the concrete number is annotated, other times the noun following is annotated as well: \textit{4 people} vs $4$. 

A similar analysis of DSTC8 is provided in Appendix~\ref{appendix:dstc8}. Given that the cutting-edge SL models are rewarded only if they provide the exact span match (see \S\ref{sec:experiments}), it seems that they get penalized mostly due to the detected annotation inconsistencies and errors in training and test data. Correcting the inconsistencies would further improve their performance, even to the point of considering the current SL benchmarks `solved' in their full-data setups. Our simple analysis thus also hints that the community should invest more effort into creating more challenging SL benchmarks in future work.

\section{Related Work}

\sparagraph{Slot Labeling in Dialog}
A variety of approaches have been proposed to leverage the semantic knowledge of PLMs like BERT \cite{devlin2019bert} and RoBERTa \cite{liu2019roberta} for intent classification and dialog state tracking \citep{chen2019bert,casanueva2020ic,louvan2020recent, gao2020machine}. The potential of the PLMs has also been exploited in end-to-end multi-domain systems, offering both design simplicity and superior performance over modular systems \citep{hosseini2020simple,peng2021soloist}.

The SL task has also benefited from the semantic prowess of PLMs. One family of models employs universal sentence encoders \cite{devlin2019bert} 
and trains a task-specific head to extract slot value spans \citep{chao2019bert,coope2020span,rastogi2020schema}.  In more recent work, \citet{henderson2021convex} define a novel SL-oriented pretraining objective. The proposed model, ConVEx, achieved substantial improvements in the SL task, particularly in low-data regimes. However, contrary to \modelname it requires training additional context-related features during fine-tuning. 
Another line of work relies on reformulating slot labeling as a natural language response generation task by adapting generative language models. \citet{madotto2020language} shows that this can be done in a zero-shot fashion by priming with task-oriented context. The GenSF model \cite{mehri2021gensf} adapts the pretrained DialoGPT model for the SL task through constrained generation. These approaches also lack contextualization and do not consider efficiency-oriented fine-tuning. 

The work closest to ours is QANLU \citep{namazifar2021language}, which also reformulates SL as a QA task, showing performance gains in low-data regimes. However, QANLU did not incorporate contextual information, did not experiment with different QA resources, nor allowed for efficient and compact fine-tuning.

\rparagraph{Efficient Methods in Dialog}
Recent dialog work is increasingly interested in the efficiency aspects of both training and fine-tuning. \citet{henderson2021convex} achieve compactness by fine-tuning only a small subset of decoding layers from the full pretrained model. As mentioned, their ConVEx framework is constrained by the particularities of their pretraining regime and cannot be easily combined with a wealth of different PLMs. 

Efficient fine-tuning with easy portability can be achieved by inserting small adapter modules inside pretrained Transformers \citep{houlsby2019parameter,pfeiffer2020adapterfusion}. Adapters make controllable response generation viable for online systems by training task-specific modules per style/topic \citep{madotto2020plug}. Through the adapters injection, \citet{wang2021task, hung2021ds} overcome the dialog entity inconsistency while achieving an advantageous computational footprint, rendering adapters particularly suitable for multi-domain specialization. However, QASL is the first example of the successful incorporation of adapters to the SL task, and also with an extra focus on the most challenging low-data scenarios.

\section{Conclusion}
We have demonstrated that reformulating slot labeling (SL) for dialog as a question answering (QA) task is a viable and effective approach to the SL task. Our comprehensive evaluations over two standard SL benchmarks have validated the effectiveness and robustness of the proposed \modelname approach, yielding improvements over state-of-the-art SL models, especially in the most challenging, few-data setups. \modelname is a very versatile framework, which can profit both from manually created and automatically created QA resources, and is applicable to an array of pretrained language models. Finally, we have shown how to efficiently fine-tune effective domain-specific SL models.


\newpage
\bibliography{anthology,custom}

\begin{thebibliography}{47}
\expandafter\ifx\csname natexlab\endcsname\relax\def\natexlab#1{#1}\fi

\bibitem[{Budzianowski et~al.(2018)Budzianowski, Wen, Tseng, Casanueva, Ultes,
  Ramadan, and Ga\v{s}i\'{c}}]{Budzianowski:2018emnlp}
Pawe{\l} Budzianowski, Tsung-Hsien Wen, Bo-Hsiang Tseng, I{\~{n}}igo Casanueva,
  Stefan Ultes, Osman Ramadan, and Milica Ga\v{s}i\'{c}. 2018.
\newblock \href {http://aclweb.org/anthology/D18-1547} {{MultiWOZ - A}
  large-scale multi-domain wizard-of-oz dataset for task-oriented dialogue
  modelling}.
\newblock In \emph{Proceedings of EMNLP 2018}, pages 5016--5026.

\bibitem[{Casanueva et~al.(2020)Casanueva, Tem\v{c}inas, Gerz, Henderson, and
  Vuli\'{c}}]{casanueva2020ic}
I{\~n}igo Casanueva, Tadas Tem\v{c}inas, Daniela Gerz, Matthew Henderson, and
  Ivan Vuli\'{c}. 2020.
\newblock \href {https://aclanthology.org/2020.nlp4convai-1.5.pdf} {Efficient
  intent detection with dual sentence encoders}.
\newblock In \emph{Proceedings of the 2nd Workshop on Natural Language
  Processing for Conversational AI}, pages 38--45.

\bibitem[{Chao and Lane(2019)}]{chao2019bert}
Guan-Lin Chao and Ian Lane. 2019.
\newblock \href {https://arxiv.org/abs/1907.03040} {{BERT-DST: S}calable
  end-to-end dialogue state tracking with bidirectional encoder representations
  from transformer}.
\newblock \emph{Proceedings of Interspeech 2019}, pages 1468--1472.

\bibitem[{Chen et~al.(2019)Chen, Zhuo, and Wang}]{chen2019bert}
Qian Chen, Zhu Zhuo, and Wen Wang. 2019.
\newblock \href {http://arxiv.org/abs/1902.10909} {{BERT} for joint intent
  classification and slot filling}.
\newblock \emph{CoRR}, abs/1902.10909.

\bibitem[{Clark et~al.(2020)Clark, Luong, Le, and Manning}]{clark2020electra}
Kevin Clark, Minh-Thang Luong, Quoc~V Le, and Christopher~D. Manning. 2020.
\newblock \href {https://arxiv.org/abs/2003.10555} {Electra: {P}re-training
  text encoders as discriminators rather than generators}.
\newblock In \emph{Proceedings of ICLR 2020}.

\bibitem[{Coope et~al.(2020)Coope, Farghly, Gerz, Vuli{\'c}, and
  Henderson}]{coope2020span}
Samuel Coope, Tyler Farghly, Daniela Gerz, Ivan Vuli{\'c}, and Matthew
  Henderson. 2020.
\newblock \href {https://aclanthology.org/2020.acl-main.11} {Span-{ConveRT:
  F}ew-shot span extraction for dialog with pretrained conversational
  representations}.
\newblock In \emph{Proceedings of ACL 2020}, pages 107--121.

\bibitem[{Devlin et~al.(2019)Devlin, Chang, Lee, and
  Toutanova}]{devlin2019bert}
Jacob Devlin, Ming{-}Wei Chang, Kenton Lee, and Kristina Toutanova. 2019.
\newblock \href {https://www.aclweb.org/anthology/N19-1423} {{BERT:
  P}re-training of deep bidirectional transformers for language understanding}.
\newblock In \emph{Proceedings of NAACL-HLT 2019}, pages 4171--4186.

\bibitem[{Fisch et~al.(2019)Fisch, Talmor, Jia, Seo, Choi, and
  Chen}]{fisch2019mrqa}
Adam Fisch, Alon Talmor, Robin Jia, Minjoon Seo, Eunsol Choi, and Danqi Chen.
  2019.
\newblock \href {https://aclanthology.org/D19-5801} {{MRQA} 2019 shared task:
  Evaluating generalization in reading comprehension}.
\newblock In \emph{Proceedings of the 2nd Workshop on Machine Reading for
  Question Answering}, pages 1--13.

\bibitem[{Gao et~al.(2020)Gao, Agarwal, Jin, Chung, and
  Hakkani-Tur}]{gao2020machine}
Shuyang Gao, Sanchit Agarwal, Di~Jin, Tagyoung Chung, and Dilek Hakkani-Tur.
  2020.
\newblock \href {https://aclanthology.org/2020.nlp4convai-1.10} {From machine
  reading comprehension to dialogue state tracking: Bridging the gap}.
\newblock In \emph{Proceedings of the 2nd Workshop on Natural Language
  Processing for Conversational AI}, pages 79--89.

\bibitem[{Gao et~al.(2019)Gao, Sethi, Agarwal, Chung, and
  Hakkani-Tur}]{gao2019dialog}
Shuyang Gao, Abhishek Sethi, Sanchit Agarwal, Tagyoung Chung, and Dilek
  Hakkani-Tur. 2019.
\newblock \href {https://aclanthology.org/W19-5932} {Dialog state tracking: A
  neural reading comprehension approach}.
\newblock In \emph{Proceedings of SIGDIAL 2019}, pages 264--273.

\bibitem[{Gao et~al.(2021)Gao, Fisch, and Chen}]{Gao:2021acl}
Tianyu Gao, Adam Fisch, and Danqi Chen. 2021.
\newblock \href {https://aclanthology.org/2021.acl-long.295} {Making
  pre-trained language models better few-shot learners}.
\newblock In \emph{Proceedings of ACL-IJCNLP 2021}, pages 3816--3830.

\bibitem[{Heck et~al.(2020)Heck, van Niekerk, Lubis, Geishauser, Lin, Moresi,
  and Gasic}]{heck2020trippy}
Michael Heck, Carel van Niekerk, Nurul Lubis, Christian Geishauser, Hsien-Chin
  Lin, Marco Moresi, and Milica Gasic. 2020.
\newblock \href {https://aclanthology.org/2020.sigdial-1.4} {{T}rip{P}y: A
  triple copy strategy for value independent neural dialog state tracking}.
\newblock In \emph{Proceedings of SIGDIAL 2020}, pages 35--44.

\bibitem[{Henderson and Vuli{\'c}(2021)}]{henderson2021convex}
Matthew Henderson and Ivan Vuli{\'c}. 2021.
\newblock \href {https://aclanthology.org/2021.naacl-main.264} {{ConVEx}:
  Data-efficient and few-shot slot labeling}.
\newblock In \emph{Proceedings of NAACL-HLT 2021}, pages 3375--3389.

\bibitem[{Hosseini{-}Asl et~al.(2020)Hosseini{-}Asl, McCann, Wu, Yavuz, and
  Socher}]{hosseini2020simple}
Ehsan Hosseini{-}Asl, Bryan McCann, Chien{-}Sheng Wu, Semih Yavuz, and Richard
  Socher. 2020.
\newblock \href
  {https://proceedings.neurips.cc/paper/2020/hash/e946209592563be0f01c844ab2170f0c-Abstract.html}
  {A simple language model for task-oriented dialogue}.
\newblock In \emph{Proceedings of NeurIPS 2020}.

\bibitem[{Hou et~al.(2020)Hou, Che, Lai, Zhou, Liu, Liu, and Liu}]{hou2020few}
Yutai Hou, Wanxiang Che, Yongkui Lai, Zhihan Zhou, Yijia Liu, Han Liu, and Ting
  Liu. 2020.
\newblock \href
  {https://atmahou.github.io/attachments/atma's_acl2020_FewShot.pdf} {Few-shot
  slot tagging with collapsed dependency transfer and label-enhanced
  task-adaptive projection network}.
\newblock In \emph{Proceedings of ACL 2020}, pages 1381--1393.

\bibitem[{Houlsby et~al.(2019)Houlsby, Giurgiu, Jastrzebski, Morrone,
  De~Laroussilhe, Gesmundo, Attariyan, and Gelly}]{houlsby2019parameter}
Neil Houlsby, Andrei Giurgiu, Stanislaw Jastrzebski, Bruna Morrone, Quentin
  De~Laroussilhe, Andrea Gesmundo, Mona Attariyan, and Sylvain Gelly. 2019.
\newblock \href {https://proceedings.mlr.press/v97/houlsby19a.html}
  {Parameter-efficient transfer learning for {NLP}}.
\newblock In \emph{Proceedings of ICML 2019}, pages 2790--2799.

\bibitem[{Hung et~al.(2021)Hung, Lauscher, Ponzetto, and
  Glava\v{s}}]{hung2021ds}
Chia{-}Chien Hung, Anne Lauscher, Simone~Paolo Ponzetto, and Goran Glava\v{s}.
  2021.
\newblock \href {https://arxiv.org/abs/2110.08395} {{DS-TOD:} efficient domain
  specialization for task oriented dialog}.
\newblock \emph{CoRR}, abs/2110.08395.

\bibitem[{Kingma and Ba(2015)}]{kingma2014adam}
Diederik~P. Kingma and Jimmy Ba. 2015.
\newblock \href {http://arxiv.org/abs/1412.6980} {Adam: {A} method for
  stochastic optimization}.
\newblock In \emph{Proceedings of ICLR 2015}.

\bibitem[{Lan et~al.(2020)Lan, Chen, Goodman, Gimpel, Sharma, and
  Soricut}]{lan2019albert}
Zhenzhong Lan, Mingda Chen, Sebastian Goodman, Kevin Gimpel, Piyush Sharma, and
  Radu Soricut. 2020.
\newblock \href {http://arxiv.org/abs/1909.11942} {{ALBERT:} {A} lite {BERT}
  for self-supervised learning of language representations}.
\newblock In \emph{Proceedings of ICLR 2020}, volume abs/1909.11942.

\bibitem[{Lewis et~al.(2021)Lewis, Wu, Liu, Minervini, K{\"{u}}ttler, Piktus,
  Stenetorp, and Riedel}]{lewis2021paq}
Patrick Lewis, Yuxiang Wu, Linqing Liu, Pasquale Minervini, Heinrich
  K{\"{u}}ttler, Aleksandra Piktus, Pontus Stenetorp, and Sebastian Riedel.
  2021.
\newblock \href {https://arxiv.org/abs/2102.07033} {{PAQ:} 65 million
  probably-asked questions and what you can do with them}.
\newblock \emph{CoRR}, abs/2102.07033.

\bibitem[{Liu et~al.(2021)Liu, Yuan, Fu, Jiang, Hayashi, and
  Neubig}]{Liu:2021survey}
Pengfei Liu, Weizhe Yuan, Jinlan Fu, Zhengbao Jiang, Hiroaki Hayashi, and
  Graham Neubig. 2021.
\newblock \href {https://arxiv.org/abs/2107.13586} {Pre-train, prompt, and
  predict: {A} systematic survey of prompting methods in {Natural Language
  Processing}}.
\newblock \emph{CoRR}, abs/2107.13586.

\bibitem[{Liu et~al.(2019)Liu, Ott, Goyal, Du, Joshi, Chen, Levy, Lewis,
  Zettlemoyer, and Stoyanov}]{liu2019roberta}
Yinhan Liu, Myle Ott, Naman Goyal, Jingfei Du, Mandar Joshi, Danqi Chen, Omer
  Levy, Mike Lewis, Luke Zettlemoyer, and Veselin Stoyanov. 2019.
\newblock \href {http://arxiv.org/abs/1907.11692} {{RoBERTa:} {A} robustly
  optimized {BERT} pretraining approach}.
\newblock \emph{CoRR}, abs/1907.11692.

\bibitem[{Liu et~al.(2020)Liu, Winata, Xu, and Fung}]{liu2020coach}
Zihan Liu, Genta~Indra Winata, Peng Xu, and Pascale Fung. 2020.
\newblock \href {https://aclanthology.org/2020.acl-main.3} {{C}oach: A
  coarse-to-fine approach for cross-domain slot filling}.
\newblock In \emph{Proceedings of ACL 2020}, pages 19--25.

\bibitem[{Louvan and Magnini(2020)}]{louvan2020recent}
Samuel Louvan and Bernardo Magnini. 2020.
\newblock \href {https://doi.org/10.18653/v1/2020.coling-main.42} {Recent
  neural methods on slot filling and intent classification for task-oriented
  dialogue systems: A survey}.
\newblock In \emph{Proceedings of COLING 2020}, pages 480--496.

\bibitem[{Madotto et~al.(2020{\natexlab{a}})Madotto, Ishii, Lin, Dathathri, and
  Fung}]{madotto2020plug}
Andrea Madotto, Etsuko Ishii, Zhaojiang Lin, Sumanth Dathathri, and Pascale
  Fung. 2020{\natexlab{a}}.
\newblock \href {https://aclanthology.org/2020.findings-emnlp.219}
  {Plug-and-play conversational models}.
\newblock In \emph{Findings of the Association for Computational Linguistics:
  EMNLP 2020}, pages 2422--2433.

\bibitem[{Madotto et~al.(2020{\natexlab{b}})Madotto, Liu, Lin, and
  Fung}]{madotto2020language}
Andrea Madotto, Zihan Liu, Zhaojiang Lin, and Pascale Fung. 2020{\natexlab{b}}.
\newblock \href {https://arxiv.org/abs/2008.06239} {Language models as few-shot
  learner for task-oriented dialogue systems}.
\newblock \emph{CoRR}, abs/2008.06239.

\bibitem[{McCann et~al.(2018)McCann, Keskar, Xiong, and
  Socher}]{mccann2018natural}
Bryan McCann, Nitish~Shirish Keskar, Caiming Xiong, and Richard Socher. 2018.
\newblock \href {http://arxiv.org/abs/1806.08730} {The natural language
  decathlon: Multitask learning as question answering}.
\newblock \emph{CoRR}, abs/1806.08730.

\bibitem[{Mehri et~al.(2020)Mehri, Eric, and
  Hakkani{-}T{\"{u}}r}]{mehri2020dialoglue}
Shikib Mehri, Mihail Eric, and Dilek Hakkani{-}T{\"{u}}r. 2020.
\newblock \href {https://arxiv.org/abs/2009.13570} {{DialoGLUE:} {A} natural
  language understanding benchmark for task-oriented dialogue}.
\newblock \emph{CoRR}, abs/2009.13570.

\bibitem[{Mehri and Esk{\'{e}}nazi(2021)}]{mehri2021gensf}
Shikib Mehri and Maxine Esk{\'{e}}nazi. 2021.
\newblock \href {https://aclanthology.org/2021.sigdial-1.51} {{GenSF:
  S}imultaneous adaptation of generative pre-trained models and slot filling}.
\newblock In \emph{Proceedings of SIGDIAL 2021}, pages 489--498.

\bibitem[{Namazifar et~al.(2021)Namazifar, Papangelis, Tur, and
  Hakkani-T{\"u}r}]{namazifar2021language}
Mahdi Namazifar, Alexandros Papangelis, Gokhan Tur, and Dilek Hakkani-T{\"u}r.
  2021.
\newblock \href {https://arxiv.org/pdf/2011.03023.pdf} {Language model is all
  you need: Natural language understanding as question answering}.
\newblock In \emph{Proceedings of ICASSP 2021}, pages 7803--7807.

\bibitem[{Peng et~al.(2021)Peng, Li, Li, Shayandeh, Liden, and
  Gao}]{peng2021soloist}
Baolin Peng, Chunyuan Li, Jinchao Li, Shahin Shayandeh, Lars Liden, and
  Jianfeng Gao. 2021.
\newblock Soloist: Buildingtask bots at scale with transfer learning and
  machine teaching.
\newblock \emph{Transactions of the Association for Computational Linguistics},
  9:807--824.

\bibitem[{Pfeiffer et~al.(2021)Pfeiffer, Kamath, R{\"{u}}ckĺe, Kyunghyun, and
  Gurevych}]{pfeiffer2020adapterfusion}
Jonas Pfeiffer, Aishwarya Kamath, Andreas R{\"{u}}ckĺe, Cho Kyunghyun, and
  Iryna Gurevych. 2021.
\newblock \href {https://arxiv.org/abs/2005.00247} {{AdapterFusion:
  Non-destructive task composition for transfer learning}}.
\newblock In \emph{Proceedings of EACL 2021}.

\bibitem[{Qiu et~al.(2020)Qiu, Sun, Xu, Shao, Dai, and Huang}]{Qiu:2020survey}
Xipeng Qiu, Tianxiang Sun, Yige Xu, Yunfan Shao, Ning Dai, and Xuanjing Huang.
  2020.
\newblock \href {https://arxiv.org/abs/2003.08271} {Pre-trained models for
  {Natural Language Processing:} {A} survey}.
\newblock \emph{CoRR}, abs/2003.08271.

\bibitem[{Rajpurkar et~al.(2018)Rajpurkar, Jia, and Liang}]{rajpurkar2018know}
Pranav Rajpurkar, Robin Jia, and Percy Liang. 2018.
\newblock \href {https://aclanthology.org/P18-2124} {Know what you don{'}t
  know: Unanswerable questions for {SQ}u{AD}}.
\newblock In \emph{Proceedings of ACL 2018}, pages 784--789.

\bibitem[{Rastogi et~al.(2020)Rastogi, Zang, Sunkara, Gupta, and
  Khaitan}]{rastogi2020schema}
Abhinav Rastogi, Xiaoxue Zang, Srinivas Sunkara, Raghav Gupta, and Pranav
  Khaitan. 2020.
\newblock \href {https://aaai.org/ojs/index.php/AAAI/article/view/6394}
  {Towards scalable multi-domain conversational agents: The schema-guided
  dialogue dataset}.
\newblock In \emph{Proceedings of AAAI 2020}, pages 8689--8696.

\bibitem[{Raux et~al.(2005)Raux, Langner, Bohus, Black, and
  Eskenazi}]{raux2005let}
Antoine Raux, Brian Langner, Dan Bohus, Alan~W Black, and Maxine Eskenazi.
  2005.
\newblock Let's go public! {T}aking a spoken dialog system to the real world.
\newblock In \emph{Ninth European conference on speech communication and
  technology}.

\bibitem[{Razumovskaia et~al.(2021)Razumovskaia, Glava\v{s}, Majewska,
  Korhonen, and Vuli\'{c}}]{Razumovskaia:2021survey}
Evgeniia Razumovskaia, Goran Glava\v{s}, Olga Majewska, Anna Korhonen, and Ivan
  Vuli\'{c}. 2021.
\newblock \href {https://arxiv.org/abs/2104.08570} {Crossing the conversational
  chasm: {A} primer on multilingual task-oriented dialogue systems}.
\newblock \emph{CoRR}, abs/2104.08570.

\bibitem[{Ruder(2021)}]{Ruder:2021blog}
Sebastian Ruder. 2021.
\newblock \href {http://ruder.io/recent-advances-lm-fine-tuning} {Recent
  advances in language model fine-tuning}.

\bibitem[{Sanh et~al.(2019)Sanh, Debut, Chaumond, and
  Wolf}]{sanh2019distilbert}
Victor Sanh, Lysandre Debut, Julien Chaumond, and Thomas Wolf. 2019.
\newblock \href {http://arxiv.org/abs/1910.01108} {Distilbert, a distilled
  version of {BERT:} smaller, faster, cheaper and lighter}.
\newblock \emph{CoRR}, abs/1910.01108.

\bibitem[{Vaswani et~al.(2017)Vaswani, Shazeer, Parmar, Uszkoreit, Jones,
  Gomez, Kaiser, and Polosukhin}]{vaswani2017attention}
Ashish Vaswani, Noam Shazeer, Niki Parmar, Jakob Uszkoreit, Llion Jones,
  Aidan~N Gomez, {\L}ukasz Kaiser, and Illia Polosukhin. 2017.
\newblock \href {https://arxiv.org/abs/1706.03762} {Attention is all you need}.
\newblock In \emph{Proceedings of NeurIPS 2017}, pages 5998--6008.

\bibitem[{Wang et~al.(2021)Wang, Zhang, Guo, Dai, Chen, and Luo}]{wang2021task}
Weizhi Wang, Zhirui Zhang, Junliang Guo, Yinpei Dai, Boxing Chen, and Weihua
  Luo. 2021.
\newblock \href {https://arxiv.org/pdf/2108.13679.pdf} {Task-oriented dialogue
  system as natural language generation}.
\newblock \emph{arXiv preprint arXiv:2108.13679}.

\bibitem[{Wolf et~al.(2020)Wolf, Debut, Sanh, Chaumond, Delangue, Moi, Cistac,
  Rault, Louf, Funtowicz, Davison, Shleifer, von Platen, Ma, Jernite, Plu, Xu,
  Le~Scao, Gugger, Drame, Lhoest, and Rush}]{wolf2019huggingface}
Thomas Wolf, Lysandre Debut, Victor Sanh, Julien Chaumond, Clement Delangue,
  Anthony Moi, Pierric Cistac, Tim Rault, Remi Louf, Morgan Funtowicz, Joe
  Davison, Sam Shleifer, Patrick von Platen, Clara Ma, Yacine Jernite, Julien
  Plu, Canwen Xu, Teven Le~Scao, Sylvain Gugger, Mariama Drame, Quentin Lhoest,
  and Alexander Rush. 2020.
\newblock \href {https://aclanthology.org/2020.emnlp-demos.6} {Transformers:
  State-of-the-art natural language processing}.
\newblock In \emph{Proceedings of EMNLP 2020: System Demonstrations}, pages
  38--45.

\bibitem[{Young(2002)}]{young2002talking}
Steve Young. 2002.
\newblock Talking to machines (statistically speaking).
\newblock In \emph{Seventh International Conference on Spoken Language
  Processing}.

\bibitem[{Zaken et~al.(2021)Zaken, Ravfogel, and Goldberg}]{zaken2021bitfit}
Elad~Ben Zaken, Shauli Ravfogel, and Yoav Goldberg. 2021.
\newblock \href {https://arxiv.org/abs/2106.10199} {{BitFit: S}imple
  parameter-efficient fine-tuning for transformer-based masked
  language-models}.
\newblock \emph{CoRR}, abs/2106.10199.

\bibitem[{Zhang et~al.(2020)Zhang, Sun, Galley, Chen, Brockett, Gao, Gao, Liu,
  and Dolan}]{zhang2020dialogpt}
Yizhe Zhang, Siqi Sun, Michel Galley, Yen-Chun Chen, Chris Brockett, Xiang Gao,
  Jianfeng Gao, Jingjing Liu, and Bill Dolan. 2020.
\newblock \href {https://aclanthology.org/2020.acl-demos.30} {{DIALOGPT} :
  Large-scale generative pre-training for conversational response generation}.
\newblock In \emph{Proceedings of ACL 2020: System Demonstrations}, pages
  270--278.

\bibitem[{Zhang et~al.(2021)Zhang, Yang, and Zhao}]{zhang2020retrospective}
Zhuosheng Zhang, Junjie Yang, and Hai Zhao. 2021.
\newblock \href {https://ojs.aaai.org/index.php/AAAI/article/view/17705}
  {Retrospective reader for machine reading comprehension}.
\newblock In \emph{Proceedings of AAAI 2021}, pages 14506--14514.

\bibitem[{Zhou and Small(2019)}]{zhou2019multi}
Li~Zhou and Kevin Small. 2019.
\newblock \href {http://arxiv.org/abs/1911.06192} {Multi-domain dialogue state
  tracking as dynamic knowledge graph enhanced question answering}.
\newblock \emph{CoRR}, abs/1911.06192.

\end{thebibliography}
\bibliographystyle{acl_natbib}
\hfill

\clearpage
\appendix
\section{Statistics and Full Results on RESTAURANTS-8k and DSTC8}
\label{sec:appendix}
\begin{itemize}
\item Table~\ref{tab:data_splits} provides the exact number of examples over all slots for all the training data splits in RESTAURANTS-8k and DSTC8.
\item Table~\ref{tab:r8k} gives the exact scores related to Figure~\ref{fig:r8k_default} in the main paper.
\item Table~\ref{tab:dsct} provides the exact scores related to Figure~\ref{fig:dstc8} in the main paper.
\end{itemize}

\begin{table*}[!t]
{\footnotesize
\def\arraystretch{0.7}
\centering
\begin{tabularx}{\linewidth}{l lXXXX}
\toprule 
&   {\bf RESTAURANTS-8k}  & \multicolumn{4}{c}{\bf DSTC8} \\
\cmidrule(lr){3-6}\\
& & {\bf Buses} & {\bf Events} & {\bf Rental Cars } & {\bf Homes} \\
\midrule 
     
1/128 & 64 & --   & -- &  -- & -- \\
1/64  & 128 & --   & --  & -- & -- \\      
1/32  & 256 & 34 & 46 & 64 & 26\\      
1/16  & 512 & 70  & 93 &129  & 54\\   
1/8  & 1024  & 141 & 187 &258 & 109 \\   
1/4  & 2049 & 283 & 374& 516 &  218  \\   
1/2  & 4099 & 566 &749 & 1032 & 437\\   
1  &  8198 & 1133 &1498 & 2064 & 874 \\   
Test  &  3731 & 377 &521 & 587 & 328 \\  
\bottomrule
\end{tabularx}
}
\caption{Statistics of the data splits extracted from the RESTAURANTS-8k and DSTC8 datasets.}
\label{tab:data_splits}
\end{table*}

\begin{table*}[]
\def\arraystretch{0.7}
\centering
\begin{tabularx}{\linewidth}{l XXXXXX}
\toprule 
& \multicolumn{1}{l}{\bf GenSF} & \multicolumn{1}{l}{\bf ConVEx} & \multicolumn{1}{l}{\bf QANLU} & \multicolumn{1}{l}{\bf RoBERTa} & \multicolumn{1}{l}{\bf RoBERTa-L} & \multicolumn{1}{l}{\bf DistilBERT} \\
\midrule 
1/128          & 72.2                    & 71.7                     & 72.9                    & 84.0                      & \textbf{84.5}                        & 72.5                         \\
1/64           & 76.1                    & 76.0                     & 83.5                    & 85.2                      & \textbf{87.2}                        & 77.2                         \\
1/32           & 82.1                    & 81.8                     & 86.9                    & \textbf{89.9}                    &\textbf{89.9}                    & 82.0                         \\
1/16           & 89.7                    & 86.4                     & 90.4                    & 91.9                      & \textbf{92.0}                        & 86.9                         \\
1/8            & 91.8                    & 90.6                     & 90.7                    & 92.2                      & \textbf{92.9}                        & 87.9                         \\
1/4            & 93.2                    & 92.5                     & 91.0                    & 94.4                      & 
\textbf{94.6}                   & 89.2                         \\
1/2            & 94.3                    & 94.1                     & 94.0                    & 95.4                      & \textbf{95.6}                        & 90.7                         \\
1              & 96.1                    & 96.0                     & 95.2                    &\textbf{96.1}                   & \textbf{96.1}                      & 91.8     \\
\bottomrule
\end{tabularx}
\caption{Average F1 scores across all slots for the evaluation on the RESTAURANTS-$8$k test set.}
\label{tab:r8k}
\end{table*}

\begin{table}[]
{\footnotesize
\centering
\def\arraystretch{0.7}
\centering
\begin{tabularx}{0.8\linewidth}{l ccc}
\toprule
& \textbf{GenSF}    & \multicolumn{1}{l}{\bf ConVEx}  & \multicolumn{1}{l}{\bf \modelname} \\
\midrule
\textbf{Buses}   &   \multicolumn{1}{l}{}   & \multicolumn{1}{l}{} & \multicolumn{1}{l}{} \\
1/32        &     & 59.20     & \textbf{92.80}   \\
1/16        &     & 75.20   & \textbf{93.30}\\
1/8         &    & 84.00   & \textbf{95.50}  \\
1/4         & \multicolumn{1}{r}{90.50} & 86.70  &\textbf{95.70}\\
1/2         & & 92.60   & \textbf{96.10}  \\
1           & \multicolumn{1}{r}{\textbf{98.10}} & 96.00  & 96.50 \\
\midrule
\textbf{Events} &   \multicolumn{1}{l}{}  & \multicolumn{1}{l}{} & \multicolumn{1}{l}{}  \\
1/32        &   & 54.00  & \textbf{76.20} \\
1/16        &   & 66.60   & \textbf{89.10} \\
1/8         &  & 82.20    & \textbf{92.70}  \\
1/4         & \multicolumn{1}{r}{91.20} & 87.20 & \textbf{95.80} \\
1/2         &     & 87.30 & \textbf{97.60}  \\
1           & \multicolumn{1}{r}{94.70} & 91.70  &\textbf{97.80}  \\
\midrule
\textbf{Rental cars} &  \multicolumn{1}{l}{} & \multicolumn{1}{l}{} & \multicolumn{1}{l}{}   \\
1/32        &   & 50.3    & \textbf{83.9} \\
1/16        &   & 60.6 & \textbf{87.0} \\
1/8         &   & 77.6 & \textbf{95.9}  \\
1/4         & 93.70 & 87.4  & \textbf{95.9}  \\
1/2         &   & 91.7    & \textbf{96.5}\\
1           & \textbf{96.90} & 92.0   & 96.3 \\
\midrule
\textbf{Homes}       &   \multicolumn{1}{l}{}  & \multicolumn{1}{l}{} & \multicolumn{1}{l}{} \\
1/32        &  & 92.0& 	\textbf{95.4}  \\
1/16        &    & 92.3  & 	\textbf{95.0} \\
1/8         &      & 94.8                     &\textbf{97.9} \\
1/4         & \multicolumn{1}{r}{86.70} & 94.5     &	98.1    \\
1/2         &      & 95.6         & \textbf{98.7}\\
1           & \multicolumn{1}{r}{93.50} & 98.3        &\textbf{	99.1} \\
\bottomrule
\end{tabularx}
\caption{Average F1 scores on the DSTC8 single-domain data sets.}
\label{tab:dsct}
}
\end{table}

\section{Brief Analysis of DSTC8}
\label{appendix:dstc8}
Performance of \modelname in the full-data scenarios already leaves little room for improvement on DSTC8 in future work. The most challenging slots are \textit{pickup date} and \textit{dropoff date} from the \textit{Rental Cars} domain. As with RESTAURANTS-8k, we again observe that some mistakes made by the SL models can be attributed to ambiguous or wrong annotations. For example, we find $2$ examples where a car is rented for a single day: whether the date is \textit{pickup date} or a \textit{dropoff date} is ambiguous.


\end{document}